\pdfoutput=1

\documentclass[11pt]{article}

\usepackage[final]{acl}

\usepackage{times}
\usepackage{latexsym}
\usepackage{amsmath}
\usepackage{booktabs}

\usepackage{xcolor}

\usepackage[T1]{fontenc}

\usepackage[utf8]{inputenc}

\usepackage{microtype}

\usepackage{inconsolata}

\usepackage{graphicx}

\usepackage{soul}            
\usepackage{booktabs}
\definecolor{lightyellow}{RGB}{255, 255, 170}
\sethlcolor{lightyellow}

\title{CoCoLex: Confidence-guided Copy-based Decoding for \\ Grounded Legal Text Generation}

\author{Santosh T.Y.S.S\textsuperscript{\textnormal 1\thanks{Santosh T.Y.S.S., along with at least one of authors at JPMorgan AI Research, conceived the idea and made substantive contributions detailed in the paper while Santosh was an intern at JPMorgan Chase.}}, Youssef Tarek Elkhayat\textsuperscript{\textnormal 1}, Oana Ichim\textsuperscript{\textnormal 2}, \\
{\bf Pranav Shetty\textsuperscript{\textnormal 3}},
{\bf Dongsheng Wang\textsuperscript{\textnormal 3}}, {\bf Zhiqiang Ma\textsuperscript{\textnormal 3}}, \\ {\bf Armineh Nourbakhsh\textsuperscript{\textnormal 3}}, 
{\bf Xiaomo Liu\textsuperscript{\textnormal 3}}\\
\textsuperscript{1}School of Computation, Information, and Technology,\\
Technical University of Munich, Germany\\
\textsuperscript{2}Graduate Institute of International and Development Studies, Geneva, Switzerland\\
\textsuperscript{3}JPMorgan AI Research
}

\begin{document}
\maketitle
\begin{abstract}
Due to their ability to process long and complex contexts, LLMs can offer key benefits to the Legal domain, but their adoption has been hindered by their tendency to generate unfaithful, ungrounded, or hallucinatory outputs. While Retrieval-Augmented Generation offers a promising solution by grounding generations in external knowledge, it offers no guarantee that the provided context will be effectively integrated. To address this, context-aware decoding strategies have been proposed to amplify the influence of relevant context, but they usually do not explicitly enforce faithfulness to the context. In this work, we introduce Confidence-guided Copy-based Decoding for Legal Text Generation (CoCoLex)—a decoding strategy that dynamically interpolates the model produced vocabulary distribution with a distribution derived based on copying from the context. CoCoLex encourages direct copying based on models' confidence, ensuring greater fidelity to the source.  Experimental results on five legal benchmarks demonstrate that CoCoLex outperforms existing context-aware decoding methods, particularly in long-form generation tasks.  
\end{abstract}

\section{Introduction}
The legal domain poses unique challenges to document-grounded language generation. Legal documents are often long, structurally complex, and prone to jargon and technical language. Additionally, tasks that are grounded in legal documents such as question answering or analysis often have strict requirements regarding accuracy and faithfulness to sources. In recent years, LLMs have revolutionized the legal domain, transforming areas such as legal education \cite{choi2023ai,jiang2024leveraging}, research \cite{livermore2023language}, compliance checking \cite{hassani2024enhancing}, and even legal practice \cite{rodgers2023technology}. 

Despite their potential, their adoption is hindered by their tendency to produce hallucinations—text inconsistent with authoritative sources such as case law, statutes, regulations, contracts, and doctrines \cite{magesh2024hallucination}.
In law, where strict adherence to authoritative sources is essential, unfaithful outputs can result in inaccurate and even harmful advice \cite{chitgopkar2024accuracy}. These issues stem from the model's inability to fully ground its outputs in verifiable knowledge, leaving them prone to generating plausible-sounding yet non-factual content \cite{el2024factuality,zmigrod-etal-2024-value,santosh2024aquaechr, nourbakhsh-etal-2025-coming}. 

To address these challenges, Retrieval-Augmented Generation (RAG) has emerged as a promising approach, equipping LLMs with external knowledge chunks to ground their outputs \cite{lewis2020retrieval,borgeaud2022improving,guu2020retrieval}. However, RAG with regular decoding still struggles to effectively integrate retrieved context \cite{hagstrom2024reality}, leading to outputs that diverge from the provided context—a critical shortcoming in high-stakes legal applications \cite{magesh2024hallucination}. While some methods aim to enhance context grounding through improved pre-training 
\cite{guu2020retrieval,lewis2020retrieval,borgeaud2022improving,izacard2023atlas}, due to the high cost of pre-training, some studies focus on inference-stage methods, such as decoding strategies for white-box models \cite{shi2023trusting,kim2024adaptive,kim2024instructive,zhao2024enhancing} or prompting for black-box models \cite{zhou2023context,byerly2024effective,li2024improving}. Most of these works prioritize correctness but 
rarely evaluate whether responses 
are faithful to the provided context. Therefore, we propose a novel decoding strategy to improve context faithfulness of white-box LLMs for legal text generation. 

Legal texts often adhere to templatized structures \cite{ghosh2023dale,nair2023exploiting} and verbatim phrases to maintain interpretive precision \cite{rossi2021verbcl}. Inspired by this characteristic, we propose Confidence-guided Copy-based Decoding for Legal Text Generation (CoCoLex), which prioritizes fidelity of the generation  by explicitly guiding the model to copy tokens from the context.
Specifically, CoCoLex uses a confidence score to dynamically balance copying and generating tokens, interpolating the model’s token distribution with a copy-based distribution. 
Note that despite its conceptual similarity to pointer generator networks \cite{see2017get}, CoCoLex is training-free, operating directly on logits during  decoding. 
Its interpolation mechanism draws inspiration from kNN-LM \cite{khandelwal2019generalization}, which retrieves from external data stores of the pre-training corpus or training examples. Prior works with KNN-LM such as \citet{khandelwal2019generalization} and  \citet{wang2023knn} have focused on language modeling and open-ended text generation, CoCoLex enhances faithfulness in context-aware generation, an underexplored area. 

Our key contributions are given below: (a)  We introduce CoCoLex, a novel decoding strategy that enhances the faithfulness of generated text by encouraging copying from the context. (b) Through experiments conducted on five legal text generation datasets using two LLMs, we demonstrate that our method, CoCoLex, not only improves correctness and ensures that generated outputs remain faithful to the source but also maintains fluency and coherence, all without increasing inference overhead, especially in long-form text generation. (c) We present an extension, CoCoLex+, which allows for copying from the entire document rather than being restricted to top-retrieved chunks. This further enhances performance by leveraging a richer and more comprehensive context. (d) We show that our method can be integrated with previous approaches, indicating that its improvements are complementary to other existing methods. 


\section{Related Work}
\noindent \textbf{LLMs and Risks for legal AI.} Lawyers are increasingly adopting AI tools to enhance their practice, from drafting contracts to conducting legal research, yielding substantial efficiency gains. As of January 2024, 41 of the top 100 largest U.S. law firms and 35\% of a broader sample of 384 firms report using generative AI tools \cite{henry2024genai, collens2024aisurvey}. In the UK, 14\% of lawyers surveyed use these tools weekly or more \cite{greenhill2024generativeai}. Despite the evident benefits, Legal AI introduces significant ethical challenges, including concerns over client confidentiality, data protection, potential biases, and the critical responsibility of lawyers to supervise and ensure the accuracy of AI-generated outputs \cite{avery2023chatgpt, harasta2024cannot, chitgopkar2024accuracy}.

\citet{dahl2024large} provide a systematic assessment of LLMs for legal tasks, categorizing hallucination types in their responses, while \citet{magesh2024hallucination} examine RAG-based tools for legal QA. Their findings indicate that although retrieval-augmented models reduce hallucinations compared to general-purpose LLMs, the issue persists. In this work, we propose a novel decoding algorithm within a retrieval-augmented generation framework to mitigate hallucinations. Particularly, our approach enhances the groundedness of model outputs by improving faithfulness to the retrieved context while ensuring relevance to legal queries.

\noindent \textbf{Retrieval Augmented Generation.} RAG retrieves relevant external information, enabling LLMs to provide factual responses. The retrieval component in RAG can operate at different granularities, such as chunk-level \cite{guu2020retrieval, lewis2020retrieval},
token-level \cite{khandelwal2019generalization, yogatama2021adaptive}, and entity-level \cite{de2021mention, fevry2020entities}. The retrieved information then can be integrated at three possible levels: (a) At the input layer, the retrieved segments can be combined with the input or query, and processed them jointly through the model \cite{ram2023context, izacard2023atlas}. 
(b) Intermediate-layer integration uses semi-parametric modules to incorporate retrieved information into the internal layers of the model \cite{wu2022memorizing, borgeaud2022improving}.
(c) Output-layer integration merges the retrieval and generation results after processing \cite{khandelwal2019generalization, santosh2024mind}.

Existing Retrieval-Augmented Generation (RAG) approaches can be categorized into training-free and training-based methods. Training-based methods involve fine-tuning both the retriever and generator, either independently \cite{karpukhin2020dense, zhou2022docprompting}, sequentially \cite{borgeaud2022improving, yoran2023making, lin2023ra, sarto2022retrieval}, or jointly \cite{rubin2024retrieval, izacard2023atlas}, allowing these components to work synergistically. In contrast, training-free methods utilize retrieved knowledge during inference by incorporating it into prompts \cite{jiang2023active, khattab2022demonstrate, trivedi2022interleaving}. These methods are computationally efficient and therefore we focus on this common variant of RAG, that performs retrieval based on the query and integrates the retrieved context within the input prompt.

However, simply augmenting the input with context may not consistently ensure alignment between the generated output and the retrieved context. To address this, decoding-based strategies \cite{shi2023trusting, kim2024adaptive, kim2024instructive, zhao2024enhancing} and prompt-based techniques \cite{zhou2023context, byerly2024effective, li2024improving} are often employed to guide the model towards generating outputs that remain faithful to the provided context. Existing decoding strategies, such as contrastive decoding \cite{li2022contrastive}, aim to amplify the influence of the retrieved context by adjusting token logits but do not explicitly enforce faithfulness to the source. To overcome this limitation, we propose a novel decoding method that incorporates a token-level copying mechanism from the retrieved context, guided by the model's confidence, in our approach.

\section{CoCoLex}

Given an input query $x$ and context $c$, a language model with parameters $\theta$ is prompted to generate a response $\mathbf{y} = \{y_1, y_2, \ldots, y_n\}$ of length $n$. The response is generated autoregressively, with each token $y_t$ sampled from the conditional probability distribution: $y_t \sim p_\theta(y_t \mid c, x, y_{<t})$. Our decoding strategy, CoCoLex consists of two major components: copy-based decoding, which derives distribution over vocabulary tokens based on copying from context and confidence factor which guides model between copying and generating. We explain these two components below. 

\subsection{Copy-based Decoding}
We hypothesize that while augmenting the query with relevant context improves response quality, it does not always guarantee that the responses are grounded in the provided context.  To address this, we introduce a copy mechanism that explicitly directs the model to copy tokens from the context during the decoding process, thereby maintaining contextual fidelity, while ensuring the response remains fluent and relevant to the query.

During decoding, we extract and store the hidden state representations of all tokens within the context \( c \) from the language model. These representations, denoted as \( h_i \) for each token, are stored along with their corresponding next token. Since the hidden states are computed during autoregressive generation, no additional forward passes are required for this storing step. At each decoding step \( t \), the model generates a hidden state vector \( h_t \), which is then compared to the stored context vectors \( \{ h_i \} \). The similarity between \( h_t \) and each context vector \( h_i \) is computed using the Euclidean \( L_2 \) distance which is subsequently transformed into similarity scores via an exponential decay (See Appendix \ref{app:ablation} for further details):
\[
s_t(i) = \exp\left(-\text{dist}_t(i)\right)
\]
where \( \text{dist}_t(i) \) represents the Euclidean distance between \( h_t \) and \( h_i \). These similarity scores are aggregated to form a probability distribution over the vocabulary. For each token \( v \) in the vocabulary, the probability of selecting \( v \) is proportional to the sum of similarity scores associated with the context tokens mapped to \( v \). The probability distribution for the copy mechanism is given by:
\[
p_\text{copy}(y_t = v \mid c, x) = \frac{\sum_{i \in \text{tokens}(v)} s_t(i)}{\sum_{v' \in V} \sum_{i \in \text{tokens}(v')} s_t(i)},
\]
where \( \text{tokens}(v) \) represents the context tokens that map to the token \( v \), and \( V \) is the vocabulary.

To reduce computational overhead, we limit the aggregation to the top-\( k \) most similar context vectors. This approximation is justified because tokens with low similarity scores contribute negligibly to the overall probability distribution. Thus, we compute copy distribution normalizing over the restricted subset of the vocabulary \( V_k \), formed by the top-\( k \) tokens,
\[
p_\text{copy}(y_t = v \mid c, x) \propto \sum_{i \in \text{top-}k \cap \text{tokens}(v)} s_t(i).
\]

\subsection{Confidence-Guidance}
Uncertainty serves as a critical metric for determining when to reliably trust the predictions of LLMs. Several prior studies have explored the use of uncertainty to detect hallucinations, highlighting its potential to identify unfaithful content \cite{vashurin2024benchmarking,fadeeva2023lm,duan2023shifting,kadavath2022language,huang2023look}. Advanced LLMs are expected to assign low probabilities to tokens that are likely to introduce inaccuracies or hallucinations \cite{kadavath2022language}. Building on this insight, we use a logit-based uncertainty approach to define a confidence indicator. Leveraging this indicator, our method dynamically balances reliance on the model's predictions and an external copy mechanism.

At each decoding step \(t\), the model produces a probability distribution over the vocabulary. The entropy of this distribution serves as a measure of uncertainty, capturing the dispersion of probabilities. Higher entropy values indicate greater uncertainty, while lower entropy values suggest that the model is more confident in its predictions:
\[
\begin{split}
H_t = -\sum_{v \in V} p_\theta(y_t = v) 
\cdot \log p_\theta(y_t = v)
\end{split}
\]
We normalize the entropy by dividing it by the maximum possible entropy, \(\log(|V|)\), which corresponds to a uniform distribution:
\[
H_t^\text{norm} = \frac{H_t}{\log(|V|)}
\]
The confidence score, \(\lambda_t\), is derived using an exponential transformation of the normalized entropy. This transformation ensures that lower entropy values correspond to higher confidence scores:
\[
\lambda_t = \exp(-H_t^\text{norm})
\]
To prevent erratic behavior caused by sudden spikes or drops in uncertainty, we smooth the confidence scores by incorporating historical data. Specifically, the value is calculated using a smoothing factor that combines its current value with a running average of values from a specified past window,
which mitigates any fluctuations and provides a more stable confidence metric. The confidence score interpolates between the model's predictions and the copy mechanism. A high confidence score increases reliance on the model’s predictions, while a low score shifts preference to the copy mechanism. The final probability distribution at decoding step \(t\) is given by:
\[
\begin{aligned}
p(y_t \mid c, x, y_{<t}) &= \lambda_t \cdot p_\theta(y_t \mid c, x, y_{<t}) \\
&\quad + (1 - \lambda_t) \cdot p_\text{copy}(y_t \mid c, x, y_{<t})
\end{aligned}
\]

\section{Experiments}
\subsection{Metrics}
We evaluate the generated answers against the reference answers for \textbf{correctness} using both lexical and semantic similarity metrics. For lexical similarity, we compute the ROUGE-L F1 score \cite{lin2004rouge}. For semantic similarity, we use AlignScore \cite{zha2023alignscore}, which computes the alignment score for each sentence in the generated answer against the reference answer and is aggregated to derive the overall score. For \textbf{faithfulness}, we assess the alignment between the generated answer and the provided source context using AlignScore \cite{zha2023alignscore}. We also evaluate stylistic properties such as \textbf{fluency} and \textbf{coherence} using UniEval \cite{zhong2022towards}, which assesses the quality of individual sentences and whether all the sentences collectively form a connected narrative.

\subsection{Datasets}
We experiment with the following datasets: \textbf{CUAD} \cite{hendrycks2021cuad} frames contract information extraction as a question-answering task, pairing contracts with relevant questions and extractive answer spans. 
\textbf{OALQA} \cite{butler-2023-open-australian-legal-dataset}  is a Question answering dataset based on the Open Australian Legal Corpus, where each question is paired with the respective reference document, along with the reference answer.  \textbf{ObliQA} \cite{gokhan2024regnlp} is a regulatory QA dataset from Abu Dhabi Global Markets financial regulations. Questions are paired with extractive clauses answering them within the whole document corpus. \textbf{AQuAECHR} is a legal question-answering dataset based on European Court of Human Rights (ECHR) judgments. Given a legal query and a corpus of ECHR judgements, the system must generate an answer following a retrieve-then-generate paradigm. 
\textbf{CLERC} \cite{hou2024clerc} focuses on legal analysis generation for U.S. federal case documents. The model generates text continuations containing legal reasoning along with citations to relevant cases based on preceding case document content, which typically introduces case facts. 


Unlike CUAD and OALQA, where each question is linked to a specific reference document, datasets like AQuAECHR, CLERC, and ObliQA require retrieval across entire document corpora. This retrieval setup is particularly challenging in legal contexts due to the ambiguity of legal queries and the interpretive complexity of legal precedents \cite{dworkin1986laws}, in contrast to traditional information-seeking queries \cite{kwiatkowski2019natural,rajpurkar2016squad} which usually have clear, unambiguous references. Effective retrieval systems in legal contexts must integrate both textual and non-textual factors, such as jurisdiction, time period, and specific conditions, to ensure the relevance and authority of retrieved content \cite{santosh2024ecthr}.
As our primary focus is on evaluating generative models' ability to produce faithfully grounded answers, we simulate these tasks within Oracle Documents—relevant documents obtained from reference answers (available as citations to documents in datasets like CLERC and AQuAECHR or verbatim extractive snippets in ObliQA), rather than across entire corpora. 
Then, a retrieval step is performed within these documents to extract relevant paragraphs or chunks, which are then provided as context for the model to synthesize answers that remain contextually grounded and relevant to the query. Detailed statistics 
of these datasets are provided in Table \ref{data-stat} of the Appendix.

\begin{table*}[!htb]
\centering
\scalebox{0.9}{
\begin{tabular}{lcccccccccc}
\toprule
& \multicolumn{5}{c}{Mistral-7B-Instruct-v0.3} & \multicolumn{5}{c}{Saul-7B-Instruct-v1} \\
\cmidrule(lr){2-6} \cmidrule(lr){7-11} 
& Cor-RL & Cor-AS & Fth-AS & Flu & Coh & Cor-RL & Cor-AS & Fth-AS & Flu & Coh \\ 
 \midrule
\multicolumn{11}{c}{CUAD}                                                                         \\ \midrule
Regular & 54.29   & 68.24   & 76.31   & \textbf{82.14*}  & 62.87 & 21.79  & 38.70  & 73.16  & \textbf{77.40}  & 70.47 \\
CAD     & 54.57   & 69.57   & 79.55   & 80.41  & 59.95 & 23.42  & 40.62  & 74.65  & 76.30  & 69.16 \\
AdaCAD  & 54.54   & 69.63   & 79.56   & 79.32  & 61.43 & 23.35  & 40.60  & 74.55  & 76.82 & 69.63 \\
CoLex   & 55.29   & 70.65   & 80.66   & 81.22  & 62.27 & 23.87  & 45.63  & 82.06  & 76.87 & 69.41 \\
CoCoLex & \textbf{55.77*}   & \textbf{71.06*}   & \textbf{80.96*}   & 81.79  & \textbf{62.73*} & \textbf{25.04*}  & \textbf{49.63*}  & \textbf{84.84*}  & 77.23 & \textbf{70.91*} \\ \midrule
\multicolumn{11}{c}{OALQA}                                                                        \\ \midrule
Regular & 40.53   & 41.39   & 59.85   & 79.78  & 84.62 & 40.16  & 32.26  & 52.84  & 79.32 & 81.94 \\
CAD     & 39.90   & 42.90   & 59.00   & 78.32  & 79.60  & 39.57  & 31.74  & 52.89  & 76.75 & 73.90  \\
AdaCAD  & 39.96   & 42.49   & 59.44   & 77.91  & 79.25 & 39.61  & 31.68  & 52.78  & 76.50  & 73.74 \\
CoLex   & 46.50   & 48.61   & 60.14   & 79.66  & 86.44 & 50.10  & 50.41  & 57.23  & 78.36 & 84.19 \\
CoCoLex & \textbf{48.34*}   & \textbf{49.84*}   & \textbf{60.87*}   & \textbf{79.94*}  & \textbf{87.24*} & \textbf{50.91*}  & \textbf{52.74*}  & \textbf{59.19*}  & \textbf{80.04*} & \textbf{86.02*} \\ \midrule
\multicolumn{11}{c}{ObliQA}                                                                       \\ \midrule
Regular & 33.86   & 73.35   & 90.84   & \textbf{75.38*}  & 70.22 & 16.90  & 62.50  & 83.10  & 72.16 & 69.95 \\
CAD     & 35.83   & 71.14   & 89.73   & 70.24  & 63.13 & 15.74  & 61.49  & 82.00  & 62.69 & 53.88 \\
AdaCAD  & 35.72   & 71.04   & 89.61   & 69.73  & 62.98 & 15.64  & 61.09  & 81.87  & 62.05 & 53.57 \\
CoLex   & 43.41   & 85.35   & 93.48   & 73.10   & 72.37 & 22.98  & 82.13  & 90.15  & 71.16 & 67.64 \\
CoCoLex & \textbf{45.12*}   & \textbf{86.01*}   & \textbf{95.96*}   & 74.41  & \textbf{74.74*} & \textbf{23.50*}  & \textbf{83.44*}  & \textbf{91.05*}  & \textbf{72.56*} & \textbf{70.18*} \\ \midrule
\multicolumn{11}{c}{AQuAECHR}                                                                     \\ \midrule
Regular & 21.77   & 52.79   & 89.66   & 74.71  & 79.99 & 17.87  & 48.30  & 80.68  & 73.82 & \textbf{68.66*} \\
CAD     & 22.04   & 49.15   & 89.28   & 72.28  & 69.43 & 18.71  & 40.60  & 80.17  & 59.82 & 61.44 \\
AdaCAD  & 22.13   & 48.69   & 89.37   & 71.27  & 69.06 & 18.68  & 40.68  & 80.58  & 59.41 & 62.35 \\
CoLex   & 29.12   & 59.79   & 91.85   & \textbf{80.76}  & 86.01 & 28.97  & 65.89  & 90.25  & 72.75 & 66.84 \\
CoCoLex & \textbf{29.84*}   & \textbf{60.10*}   & \textbf{92.27*}   & 80.48  & \textbf{86.44*} & \textbf{29.28*}  & \textbf{66.26*}  & \textbf{91.15*}  & \textbf{74.24*} & 67.91 \\ \midrule
\multicolumn{11}{c}{CLERC}                                                                        \\ \midrule
Regular & 10.42   & 42.38   & 74.02   & 77.27  & 78.41 & 9.39   & 23.40  & 55.91  & 73.76 & 65.24 \\
CAD     & 10.56   & 34.98   & 66.35   & 74.37  & 73.09 & 9.06   & 24.10  & 56.04  & 71.09 & 65.17 \\
AdaCAD  & 10.52   & 35.11   & 66.46   & 74.62  & 73.13 & 9.09   & 24.18  & 56.02  & 71.17 & 65.15 \\
CoLex   & 12.71   & 54.94   & 78.62   & \textbf{78.16*}  & 89.55 & \textbf{12.78*}  & 33.02  & 59.58  & \textbf{76.36} & 71.01 \\
CoCoLex & \textbf{12.88*}   & \textbf{58.12*}   & \textbf{79.54*}   & 77.92  & \textbf{90.55*} & 12.51  & \textbf{34.95*}  & \textbf{62.79*}  & 76.25 & \textbf{73.93*} \\ \bottomrule
\end{tabular}}
\caption{Performance comparison of different decoding-based methods across five legal text generation datasets, using two language models. Cor, Fth, RL, AS, Flu, Coh denote Correctness, Faithfulness, ROUGE-L, AlignScore, Fluency and Coherence respectively. Entries marked with * are statistically significantly higher than the second-best performing baseline at the 95\% confidence level, according to the Wilcoxon signed-rank test.}
\label{tab_main}
\end{table*}
 
\subsection{Baselines and Implementation Details}
We compare CoCoLex to the following baselines: 1) Regular Decoding. 2) CAD (Context-aware Decoding) \cite{shi2023trusting}, which enhances groundedness by sampling from a contrastive output distribution that amplifies the difference between output probabilities with and without context. CAD modifies the model's original output distribution by incorporating the pointwise mutual information (PMI) between the context \( c \) and the generation \( y_t \), conditioned on \( x, y_{<t} \):
\begin{align*}
y_t \sim \text{softmax}&\left[(1 + \alpha) \, \text{logit}_\theta(y_t \mid c, x, y_{<t}) \right. \\
   &\quad \left. - \alpha \, \text{logit}_\theta(y_t \mid x, y_{<t}) \right].
\end{align*}
A larger \(\alpha\) places more weight on the adjustment, while \(\alpha = 0\) reduces to regular decoding. 
3) AdaCAD (Adaptive CAD) \cite{wang2024adacad} dynamically infers $\alpha$ in CAD at every timestamp to get $\alpha_t$  based on the degree of conflict, measured by the Jensen-Shannon divergence (JSD) between the distributions representing contextual and parametric knowledge:
\[
\alpha_t = \text{JSD}\left(p_\theta(y_t \mid x, y_{<t}) \parallel p_\theta(y_t \mid c, x, y_{<t})\right).
\]
We also derive CoLex, from CoCoLex, removing confidence-based dynamic interpolation and interpolating based on a static value to study the effect of the confidence guidance. 

We apply these methods to \texttt{mistralai/ Mistral-7B-Instruct-v0.3} \cite{jiang2023legal} and \texttt{Equall/Saul-7B-Instruct-v1} \cite{colombo2024saullm}, with Saul being specifically pre-trained on a legal unsupervised corpus and instruction corpus, leveraging the base Mistral model. We use BM25 \cite{robertson2009probabilistic} for retrieving top-k passages from the documents due to its strong performance in legal retrieval settings \cite{santosh2024ecthr,rosa2021yes}. Implementation details are provided in App. \ref{impl}.

\subsection{Results}
We present the results on the five datasets in Table \ref{tab_main}. We observe that context-aware decoding methods such as CAD and AdaCAD improve both correctness and faithfulness over regular decoding on the CUAD dataset across both models. However, they exhibit a decline in fluency and coherence scores. On OALQA, these methods improve correctness (as measured by AlignScore) only with the Mistral model and achieve comparable faithfulness scores across both models. For other datasets, including ObliQA, AQuAECHR, and CLERC, CAD and AdaCAD lead to a decline in both correctness and faithfulness across both models, although they show some improvement in lexical-based ROUGE scores. These results underscore that CAD and AdaCAD are primarily effective for short-text generation (as seen in CUAD) but struggle with long-form generation, particularly in maintaining fluency and coherence. Among them, AdaCAD performs slightly better, especially in long-range tasks.

Our proposed CoLex outperforms prior approaches by guiding the model to explicitly copy tokens from the context. This approach not only enhances faithfulness by aligning generations more closely with the provided context but also improves correctness. The improvement is more pronounced in long-range tasks. The copying mechanism does not limit CoLex's fluency and coherence, which are comparable to regular decoding, with only a marginal decline in some datasets. CoCoLex further enhances correctness and faithfulness by dynamically balancing copying and text generation based on the model’s confidence. It also improves fluency and coherence compared to CoLex. Across both models, we observe that the legally pre-trained Saul underperforms the generalist Mistral model, consistent with prior findings \cite{santosh2024aquaechr}. This is mainly due to Saul’s difficulty in synthesizing information from the provided context when following instructions. However, our proposed methods substantially improve Saul’s performance, mitigating its limitations through copy-based decoding, which explicitly guides the model to copy relevant tokens. Overall, our approach enhances faithfulness and correctness without compromising fluency and coherence.

\begin{table}[h]
\centering
\scalebox{0.8}{
\begin{tabular}{lcccc}
\toprule
 & Cor & Fth & Flu & Coh \\
\midrule
Regular & 4.40 & 4.24 & 4.88 & 4.88 \\
AdaCAD  & 4.24 & 3.84 & 4.80 & 4.84 \\
CoCoLex & \textbf{4.64} & \textbf{4.44} & \textbf{4.96} & \textbf{4.92} \\
\bottomrule
\end{tabular}}
\caption{Human evaluation results on 25 randomly sampled questions from the AQuAECHR dataset.}
\label{tab_human_eval}
\end{table}

\noindent \textbf{Human Evaluation.} We randomly sample 25 questions from the AQuAECHR dataset and generate responses using three methods—Regular, AdaCAD, and CoCoLex—applied to the Mistral-7B model, yielding a total of 75 responses. Each response is assessed by a legal ECHR expert, across four criteria: correctness (relevance to the question and alignment with the reference answer), faithfulness (adherence to the provided passages in context), fluency, and coherence. Evaluations are conducted on a 5-point Likert scale, where 1 represents the lowest quality and 5 the highest. Table \ref{tab_human_eval} presents the average scores for each criterion across methods. Notably, the legal expert consistently ranked CoCoLex higher than Regular and AdaCAD, reinforcing its superiority in generating legally faithful responses. Additionally, the results highlight that AdaCAD underperforms compared to Regular in this task of long-range text generation, particularly struggling with faithfulness. A detailed case study is provided in Appendix \ref{case}

\subsection{Discussion and Analysis}
\subsubsection{Providing Document Context}
Retrieval-augmented approaches for handling lengthy documents typically follow a two-step process: first, retrieving relevant evidence passages or chunks from these documents and then using these retrieved passages as context for the generator to generate an answer. However, this approach often suffers from chunking-related issues, where improper segmentation and concatenation of retrieved passages disrupt the semantics, leading to incomplete and incoherent information retrieval \cite{qian2024grounding, dong2023survey}. This, in turn, makes it difficult for the model to stay aligned with the main query, eventually degrading the accuracy of the generated response.

To address this limitation, we introduce CoCoLex+, an extension of CoCoLex that leverages the entire document’s encoded hidden states rather than restricting copying only the tokens in top-retrieved passages. By incorporating representations from the full document, CoCoLex+ enables the model to capture a more comprehensive understanding of the document’s content, leading to more contextually grounded responses. We chunk documents into overlapping segments to efficiently obtain these hidden state representations and extract contextualized hidden states for each token. To prevent redundancy across overlapping contexts, each token is assigned a single hidden state representation, taken from the chunk where it has the most autoregressive context. During inference, we continue to limit explicit textual context to the top-k retrieved passages, similar to CoCoLex, but augment it with hidden states from the full document to facilitate copying.

\begin{table}[!htb]
\scalebox{0.8}{
\begin{tabular}{lcccccc}
\toprule
         & C-R & C-A & F-P & F-D  & Fl   & Co   \\  \midrule
\multicolumn{7}{c}{CLERC - Mistral-7B-Instruct-v0.3}        \\ \midrule
CoCoLex  & 12.88  & 58.12  & 79.54 & 89.34  & \textbf{77.92} & \textbf{90.55} \\ 
CocoLex+ & \textbf{13.01}  & \textbf{60.66}  & \textbf{80.17} & \textbf{90.12}  & 77.84 & 90.52 \\ \midrule
\multicolumn{7}{c}{CLERC - Saul-7B-Instruct-v1}             \\ \midrule
CoCoLex  & 12.51  & 34.95  & 62.79 & 62.71 & \textbf{76.25} & 73.93 \\
CocoLex+ & \textbf{13.33}  & \textbf{44.91}  & \textbf{72.51} & \textbf{70.31}  & 75.53 & \textbf{75.16} \\ \midrule
\multicolumn{7}{c}{AQuAECHR - Mistral-7B-Instruct-v0.3}     \\ \midrule
CoCoLex  & 29.84  & 60.10  & 92.27 & 61.35 & \textbf{80.48} & \textbf{86.44} \\
CocoLex+ & \textbf{30.06}  & \textbf{60.37}  & \textbf{92.62} & \textbf{62.71}  & 80.26 & 86.28 \\ \midrule
\multicolumn{7}{c}{AQuAECHR - Saul-7B-Instruct-v1}          \\ \midrule
CoCoLex  & 29.28  & 66.26  & \textbf{91.15} & 69.39  & 74.24 & \textbf{67.91} \\
CocoLex+ & \textbf{29.63}  & \textbf{67.45}  & 88.39 & \textbf{71.53}  & \textbf{74.29} & 67.42 \\ \bottomrule
\end{tabular}}
\caption{Performance comparison between CoCoLex and CoCoLex+, where CoCoLex+ enhances the copy mechanism to apply to all tokens in the document, rather than being limited to tokens in the retrieved chunks appended to the prompt. C-R(A), F-P(D), Fl, Co denote Correctness-ROUGE-L(AlignScore), Faithfulness-Passages (Documents), Fluency and Coherence.}
\label{tab_plus}
\end{table}

We evaluate CoCoLex+ on long-range generation tasks—CLERC and AQuAECHR—and present the results in Table \ref{tab_plus}. Given that CoCoLex+ enables copying from the entire document, we extend our faithfulness evaluation to measure alignment with the full document, in addition to the top-retrieved passages, as done previously. Our results show that CoCoLex+ consistently outperforms CoCoLex in correctness across both datasets and models.  Interestingly, it also enhances faithfulness with respect to retrieved passages on CLERC for both models and on AQuAECHR for Mistral, suggesting that additional global context improves the model’s ability to stay grounded in the provided evidence. However, in AQuAECHR with Saul, we observe a slight decline in passage-level faithfulness and an increase in document-level faithfulness, indicating that Saul benefits more from whole-document copying than from restricting itself to retrieved tokens. Crucially, CoCoLex+ maintains fluency and coherence comparable to CoCoLex, despite handling a broader copying vocabulary. These findings highlight that expanding the model’s access to full-document representations strengthens its ability to generate accurate, well-grounded responses, making CoCoLex+ particularly effective for long-form legal text generation.
 
\begin{table}[!htb]
\centering
\scalebox{0.85}{
\begin{tabular}{lccccc}
\toprule
                 & C-RL & C-AS & F-AS & Flu   & Coh   \\ \midrule
\multicolumn{6}{c}{CUAD - Mistral-7B-Instruct-v0.3}         \\ \midrule
CoCo          & 55.77  & 71.06  & 80.96  & 81.79 & \textbf{62.73} \\
Ada + CoCo & \textbf{56.26}  & \textbf{71.87}  & \textbf{81.18}  & \textbf{81.99} & 62.41 \\ \midrule
\multicolumn{6}{c}{CUAD - Saul-7B-Instruct-v1}              \\ \midrule
CoCo          & 25.04  & 49.63  & 84.84  & \textbf{77.23} & \textbf{70.91} \\
Ada + CoCo & \textbf{25.38}  & \textbf{50.46}  & \textbf{85.05}  & 77.18 & 70.89 \\ \midrule
\multicolumn{6}{c}{CLERC - Mistral-7B-Instruct-v0.3}        \\ \midrule
CoCo          & \textbf{12.88}  & \textbf{58.12}  & \textbf{79.54}  & \textbf{77.92} & \textbf{90.55} \\
Ada + CoCo & 12.10  & 50.24  & 72.09  & 75.51 & 86.51 \\ \midrule
\multicolumn{6}{c}{CLERC - Saul-7B-Instruct-v1}             \\ \midrule
CoCo          & 12.51  & 34.95  & 62.79  & \textbf{76.25} & \textbf{73.93} \\
Ada + CoCo & \textbf{13.33}  & \textbf{36.75}  & \textbf{63.30}  & 75.09 & 73.45  \\  \midrule
\multicolumn{6}{c}{AQuAECHR - Mistral-7B-Instruct-v0.3}     \\ \midrule
CoCo          & \textbf{29.84}  & \textbf{60.10}  & \textbf{92.27}  & \textbf{80.48} & \textbf{86.44} \\
Ada + CoCo & 27.74  & 55.89  & 90.70  & 80.19 & 81.84 \\ \midrule
\multicolumn{6}{c}{AQuAECHR - Saul-7B-Instruct-v1}          \\ \midrule
CoCo         & \textbf{29.28}  & \textbf{66.26}  & \textbf{91.15}  & \textbf{74.24} & \textbf{67.91} \\
Ada + CoCo & 26.30  & 57.79  & 90.10  & 69.15 & 63.71 \\
\bottomrule
\end{tabular}}
\caption{Demonstrating the complementarity of both approaches: CoCoLex combined with AdaCAD improves when AdaCAD outperforms regular decoding, but results in a decrease when AdaCAD is less effective.}
\label{tab_mix}
\end{table}

\subsubsection{Combining Strategies}
While AdaCAD grounds generation by contrastively amplifying the difference between output probabilities with and without context—down weighting prior parametric knowledge when relevant contextual information is available, CoCoLex grounds generation by guiding the model to copy from retrieved context effectively. Given their complementary mechanisms, we extend CoCoLex by incorporating AdaCAD’s contrastive probability distribution into CoCoLex’s final probability computation, modifying $p_\theta$ directly. 

We evaluate the combined AdaCAD + CoCoLex approach on CUAD, CLERC, and AQuAECHR across both models, with results in Table \ref{tab_plus}. Our findings indicate that this combination enhances performance in CUAD across both models and improves CLERC for Saul. However, we observe a decline in CLERC for Mistral and in AQuAECHR for both models. Notably, these trends closely align with AdaCAD’s performance in Table \ref{tab_main}: whenever AdaCAD improves over regular decoding (e.g., CUAD for both models, CLERC for Saul), the combined approach also yields gains. Conversely, when AdaCAD underperforms compared to regular decoding, the combination similarly results in a performance drop. This suggests that integrating AdaCAD with CoCoLex can be effective when AdaCAD itself outperforms regular decoding, reinforcing their complementarity. 

\begin{table}[!htb]
\centering
\scalebox{0.8}{
\begin{tabular}{lcc}
\toprule
           & CUAD & AQuAECHR \\ \midrule
Regular    & 1.00x                        & 1.00x                            \\
CAD        & 1.75x                    & 1.71x                         \\
AdaCAD     & 1.77x                     & 1.72x                         \\
CoLex      & 1.49x                     & 1.61x                         \\
CoCoLex    & 1.51x                     & 1.62x                         \\
Ada + CoCo & 2.31x                     & 2.25x                         \\
CoCoLex+   & 1.96x                     & 2.96x            \\ \bottomrule           
\end{tabular}}
\caption{Inference time comparison of different approaches, scaled to regular decoding.}
\label{tab_time}
\end{table}

\subsubsection{Inference Time}
We compute inference time for different methods using a randomly sampled 10\% subset of the CUAD and AQuAECHR datasets, running on an Nvidia A100 GPU. To account for variations in output length across methods, we normalize inference time by the number of tokens generated and report relative inference times compared to regular decoding (1.00x). Our findings show that CAD and AdaCAD nearly double inference time, requiring an additional decoding step at each timestep to obtain logits with and without context. In contrast, CoLex and CoCoLex, while avoiding the extra decoding step of CAD/AdaCAD, incur a smaller overhead, which is the result of: indexing hidden states of tokens in the retrieved context, retrieving, normalizing, and interpolating token probabilities with copy-based probabilities at each timestep. When combining AdaCAD and CoCoLex, these overheads accumulate linearly since their processes are non-overlapping, further increasing inference time. Additionally, CoCoLex+, with its mechanism to index hidden states from the entire document, introduces further latency—especially for longer documents in AQuAECHR, where the overhead is more pronounced than CUAD.

\section{Conclusion}
We introduced CoCoLex, a decoding strategy that enhances faithfulness in legal text generation by dynamically balancing the model’s token distribution with a copy-based distribution derived from retrieved context, guided by model confidence. Experiments on five legal datasets highlight the limitations of existing context-aware decoding methods, which improve faithfulness in short-range but struggle with long-range tasks. CoCoLex enhances generation fidelity and correctness in long-form tasks while maintaining fluency and coherence without substantial inference overhead. Our findings show that leveraging hidden state representations from the entire document, rather than restricting copying to top-retrieved chunks, mitigates context length constraints, enabling the model to capture richer information and improve faithfulness. We also find that integrating contrastive-based approaches such as AdaCAD with CoCoLex is beneficial, reinforcing their complementarity. Future work could extend copying beyond the token level to incorporate larger semantic units, such as phrases or clauses, for improved contextual alignment.

\section*{Limitations}
To evaluate the faithfulness of generator models, our experiments assume an oracle document setting, where document retrieval is performed beforehand and provided as input. However, in real-world applications requiring retrieval across an entire document corpus—such as in benchmarks like CLERC, AQuAECHR, and ObliQA—the quality and correctness of generated text inherently depend on the accuracy and completeness of retrieval. If critical legal information is missing, poorly ranked, or incorrectly retrieved, the model may still produce misleading outputs. Addressing this retrieval bottleneck remains a key challenge, particularly in legal contexts, where case law documents, regulations, and statutes can be lengthy, nuanced, and difficult to rank effectively \cite{santosh2024ecthr, locke2022case}. Unlike standard information retrieval tasks that prioritize semantic similarity, legal retrieval must account for additional constraints such as precedential value, temporal relevance \cite{santosh2024chronoslex} and procedural applicability \cite{santosh2024towards}.

While CoCoLex improves long-form generation fidelity by encouraging direct copying, it does not explicitly handle cases requiring reasoning beyond the retrieved context, such as synthesizing multiple sources, reconciling conflicting precedents, or constructing persuasive legal arguments. Moreover, as CoCoLex relies on hidden state similarities to guide copying, its effectiveness depends on the model’s ability to learn robust token representations, which may vary across architectures, pretraining objectives, and the pretraining corpus.

\section*{Ethics Statement}
All datasets used in this work are publicly available and have been utilized in compliance with their respective data usage policies. While datasets such as CLERC and AQuAECHR, which involve case law judgments, are not anonymized, our work engages with the data in a manner that we believe does not cause harm beyond the availability of this information. 

LLMs, due to the historical biases inherent in their pre-training data, may perpetuate harmful prejudices and inaccuracies, potentially exacerbating existing gaps in legal knowledge and representation. This can lead to biased outputs, including factual inaccuracies and misrepresentations of legal citations. As such, caution is necessary in the responsible deployment of LLMs for legal information-seeking tasks. While LLMs can be powerful tools, they are not intended to replace legal professionals but to assist them by augmenting their expertise. Therefore, it is critical to employ LLMs with care in legal contexts, with constant monitoring for fairness, accuracy, and alignment with legal principles.

In the responsible deployment of LLMs, there is also the need to address broader questions surrounding the automation of legal tasks. The growing use of LLMs in legal practice should be carefully assessed to ensure that the integrity, professionalism, and accountability of the legal profession are maintained. As these technologies evolve, it is important to continually evaluate the impact of LLMs on the legal system, ensuring they complement legal professionals and enhance the equitable and effective delivery of legal services. This includes ongoing reflection on their potential biases, transparency, and the ethical implications of their integration into the legal workflow.

\section*{Dislaimer}
This paper was prepared for informational purposes by the Artificial Intelligence Research group of JPMorgan Chase \& Co. and its affiliates (``JP Morgan'') and is not a product of the Research Department of JP Morgan. JP Morgan makes no representation and warranty whatsoever and disclaims all liability, for the completeness, accuracy, or reliability of the information contained herein. This document is not intended as investment research or investment advice, or a recommendation, offer or solicitation for the purchase or sale of any security,financial instrument, financial product, or service, or to be used in any way for evaluating the merits of participating in any transaction, and shall not constitute a solicitation under any jurisdiction or to any person, if such solicitation under such jurisdiction or to such person would be unlawful.

\bibliography{custom}

\appendix

\section{Dataset}
For computing metrics, in the case of CUAD and ObliQA, where the reference answers are extractive phrases or sentences from the dataset, we concatenate them to form the reference answer. 
In CUAD, which also contains questions where answers are unavailable in the contract, we focus only on instances with answers in the contract for faithfulness evaluation. Detailed stats on the dataset is provided in Table \ref{data-stat}.

\begin{table*}[]
\centering
\begin{tabular}{lcccc}
\toprule
\textbf{Dataset}  & \#Instances & \# Docs & Docs Length & Answer length \\ \midrule
\textbf{CUAD}     & 4,182    & 1.00    & 7,125.60           & 31.11         \\
\textbf{OALQA}    & 2,024    & 1.00    & 5,973.99           & 90.07         \\
\textbf{ObliQA}   & 2,786    & 1.32    & 48,138.35          & 117.55        \\
\textbf{AQuAECHR} & 1,116    & 3.11    & 14,041.33          & 193.58        \\
\textbf{CLERC}    & 1,000    & 2.79    & 4,687.04           & 187.53        \\ \bottomrule
\end{tabular}
\caption{Detailed Statistics of datasets. Docs represent oracle reference documents associated with each question. Length is obtained in terms of the number of words. We report the mean for \# Docs, Docs length and Answer length.}
\label{data-stat}
\end{table*}

\section{Implementation Details}
\label{impl}
We use greedy decoding with a Repetition Penalty of 1.5, for all our experiments. For CAD, we use a static value of $\alpha$ as 0.5. Following \cite{wang2024adacad}, we clamp the $\alpha$ value to a minimum of 0.3 in AdaCAD.  In CoCoLex and CoLex, we use the last layer to extract the hidden states for the tokens. We use a smoothing parameter of 0.5 between the prior and the current $\lambda$ value. We use 0.5 as $\lambda$ in case of CoLex. We clamp the value of $\lambda$ in [0.2,0.8] for CoCoLex. For efficient computation, we leverage the FAISS \cite{johnson2019billion} library, designed for fast nearest-neighbor retrieval in high-dimensional spaces. We retrieve the top 3 passages from the oracle documents using BM25 as context for all datasets except for ObliQA and CUAD where we used 10.

\begin{table}[!htb]
\centering
\scalebox{0.9}{
\begin{tabular}{lccccc}
\toprule
        & \#Psg & Cor-AS & Fth-AS & Flu   & Coh   \\  \midrule
Regular & 3          & 42.38  & 74.02  & 77.27 & 78.41 \\
        & 6          & 40.78  & 69.22  & 76.84 & 76.17 \\
        & 10         & 35.70  & 66.12  & 75.02 & 75.05 \\  \midrule
AdaCAD  & 3          & 35.11  & 66.46  & 74.62 & 73.13 \\
        & 6          & 31.03  & 60.77  & 73.03 & 71.23 \\
        & 10         & 29.01  & 57.50  & 71.37 & 70.91 \\
        \midrule
CoCoLex & 3          & \textbf{58.12}  & \textbf{79.54}  & \textbf{77.92} & \textbf{90.55} \\
& 6          & 57.46  & 78.48  & 76.88 & 90.01 \\
        & 10          & 56.95  & 77.80  & 76.18 & 89.31 \\
        \bottomrule
\end{tabular}}
\caption{Performance comparison of Regular, AdaCAD, and CoCoLex on the CLERC dataset using the \texttt{Mistral-7B-Instruct-v0.3} model with varying numbers of passages as context. Cor, Fth, AS, Flu, and Coh denote Correctness, Faithfulness, AlignScore, Fluency, and Coherence, respectively.}
\label{tab_num_para}
\end{table}

\begin{table}[]
\centering
\scalebox{1.0}{
\begin{tabular}{lcccc}
\toprule
Similarity         & Cor-AS & Fth-AS & Flu   & Coh   \\ \midrule
Cosine    & 55.16  & 77.26  & 76.42 & 87.28 \\
Euclidean & \textbf{58.12}  & \textbf{79.54}  & \textbf{77.92} & \textbf{90.55} \\ \bottomrule
\end{tabular}}
\caption{Performance comparison of CoCoLex on the CLERC dataset using the \texttt{Mistral-7B-Instruct-v0.3} model with different distance metrics for computing similarity.}
\label{tab_distance}
\end{table}

\begin{table}[]
\centering
\scalebox{1.0}{
\begin{tabular}{lcccc}
\toprule
Layer & Cor-AS & Fth-AS & Flu   & Coh   \\ \midrule
-1                   & \textbf{58.12}  & \textbf{79.54}  & \textbf{77.92} & \textbf{90.55} \\
-5                   & 54.67  & 79.01  & 72.18 & 86.12 \\
-10                  & 52.19  & 78.45  & 72.87 & 85.19 \\
-15                  & 49.79  & 78.81  & 70.29 & 82.26 \\
-20                  & 47.78  & 77.18  & 68.18 & 83.19 \\
-25                  & 46.92  & 77.29  & 65.28 & 79.68 \\ \bottomrule
\end{tabular}}
\caption{Performance comparison of CoCoLex on the CLERC dataset using the \texttt{Mistral-7B-Instruct-v0.3} model with different layers for obtaining hidden state representation for computing similarity.}
\label{tab_layer}
\end{table}

\section{Ablation Study}
\label{app:ablation}
\noindent \textbf{Number of Passages in the context} We vary the number of top passages retrieved using BM25 from the oracle documents and append them as context. Table \ref{tab_num_para} reports the performance of Regular, AdaCAD, and CoCoLex using the \texttt{Mistral-7B-Instruct-v0.3} model on the CLERC dataset. As the number of retrieved passages increases, more distractors are introduced, and performance scores consistently decline across all methods. Notably, Regular decoding experiences a sharp drop in correctness and faithfulness, highlighting its brittleness to distractors. While AdaCAD is more robust than Regular due to its contrastive decoding mechanism, which enhances reliance on context, it also becomes susceptible to distractors. In contrast, CoCoLex leverages confidence and similarity values to assess the relevance of retrieved passages, allowing it to selectively incorporate useful information based on confidence, eventually leading to filtering out distractions. This enables CoCoLex to maintain superior robustness against distractors.

\noindent \textbf{Similarity Function} We compare different similarity functions used in CoCoLex, to compute the similarity between the current hidden state and the hidden states of context tokens. Table \ref{tab_distance} reports the performance on the CLERC dataset using the \texttt{Mistral-7B-Instruct-v0.3} model. We observe that Euclidean distance consistently outperforms cosine similarity across all metrics.

\noindent \textbf{Layer for hidden state representation} We analyze the impact of selecting different layers for extracting hidden state representations of the current token and context tokens in CoCoLex. Table \ref{tab_distance} reports the performance on the CLERC dataset using the \texttt{Mistral-7B-Instruct-v0.3} model across different layers. Results indicate that using the last layer is more effective, as it provides a refined and contextually informed representation of the token, incorporating the full extent of model reasoning. In contrast, earlier layers primarily encode lower-level features and intermediate transformations, which may not fully capture the semantic and contextual nuances necessary for robust similarity computation. As a result, they are less effective in guiding accurate token retrieval for copying from the source context in CoCoLex.

\begin{table*}[h]
\centering
\scalebox{0.8}{
\begin{tabular}{p{1.5cm}p{18cm}}
\toprule
\toprule
Question & In the context of restrictions on the right to marry, how does the Court determine whether the restrictions are properly regulated and subjected to judicial review, and how does this impact the applicant's ability to challenge the denial of authorization? \\
\midrule
Source Passages from 

001-96453 & 80: \hl{Article 12, they added, did not guarantee an unlimited right to marry since this right was regulated by “the national laws” governing its exercise.} Consequently, as the Court had held in the case of B. and L. v. the (no. 36536/02, judgment of 13 September 2005), this right was subject to limitations, although they could not restrict or reduce the right to such an extent that its very essence was impaired.

84: It was natural that, as the Court had held in many cases concerning the rights of prisoners, any measure depriving a person of liberty inevitably entailed limitations on the exercise of Convention rights, including a measure of control on prisoners’ contacts with the outside world. In the context of the right to marry this might mean that the authorities, in exercising their power in this area, could monitor the wedding ceremony and limit, for instance, the number of participants. \hl{However, they should maintain a fair balance between the demands of security in prison and the prisoner’s right to marry. Their discretion should be limited to, and their decisions based on, concrete facts, not on prejudice}. A refusal should be restricted to situations where the marriage ceremony would jeopardise prison security – and not just be difficult to organise. Moreover, rules regulating the authorities’ discretion should be laid down in the national law. In particular, the law ought to list specific circumstances in which the authorities should not give leave to marry – for example, if it would affect the process of rehabilitation. In this context, it must be stressed that the issues involved were of a sensitive nature; thus, unjustified refusal of leave to marry could be regarded as additional or disciplinary punishment.

90: This conclusion is reinforced by the wording of Article 12. In contrast to Article 8 of the Convention, which sets forth the right to respect for private and family life, and with which the right “to marry and to found a family” has a close affinity, Article 12 does not include any permissible grounds for an interference by the State that can be imposed under paragraph 2 of Article 8 “in accordance with the law” and as being “necessary in a democratic society”, for such purposes as, for instance, “the protection of health or morals” or “the protection of the rights and freedoms of others”. \hl{Accordingly, in examining a case under Article 12 the Court would not apply the tests of “necessity” or “pressing social need” which are used in the context of Article 8 but would have to determine whether, regard being had to the State’s margin of appreciation, the impugned interference was arbitrary or disproportionate (see paragraph 88 above, with references to the Court’s case-law).} \\ \midrule
\midrule
Regular & In the context of restrictions on the right to marry, as established by [001-96453], the Court determines whether such limitations are properly regulated and subjected to judicial review based upon two key factors. Firstly, any measures restricting this fundamental freedom must be limited to \hl{concrete facts rather than prejudice ([001-96453])}. Secondly, rules governing these discretionary decisions should be laid down in national law with specific circumstances listed where authorization for marriage may not be granted - e.g., if it would affect rehabilitation processes or jeopardize prison security (ibid.). This approach allows applicants an opportunity to challenge denials through a legal framework that ensures fairness while maintaining balance between individual rights and societal needs within democratic societies' margins of appreciation.[001-96453]
 \\ \midrule
AdaCAD & In the context of restrictions on the right to marry, [001-96453] states that Article 12 does not guarantee an unlimited right since it is regulated by "the national laws" governing its exercise. Consequently, for a restriction to be proper and subjected to judicial review, rules should govern authorities' discretion in this area with specific circumstances listed where leave to marry ought not given - such as affecting rehabilitation process ([ibid]). \hl{This means prisoners have the ability to challenge denial authorization if concrete facts were disregarded or prejudice influenced decisions rather than prison security jeopardy being involved}; moreover, refusal cannot restrict so much that essence impaired due to sensitive nature which could constitute additional punishment without justification.[001-96453]. \\ \midrule
CoCoLex & In the context of restrictions on the right to marry, the Court determines whether the restrictions are properly regulated and subjected to judicial review by assessing whether \hl{the national laws} governing the exercise of this right are in place [001-96453]. The Court also ensures that the discretion of the authorities in exercising their power to regulate the right to marry is limited to \hl{concrete facts, not prejudice}, and that their decisions are based on specific circumstances listed in the national law [001-96453]. This means that the applicant has the ability to challenge the denial of authorization if it can be \hl{shown that the restrictions imposed are arbitrary or disproportionate, as the Court would not apply the tests of "necessity" or "pressing social need" but would determine whether the impugned interference was arbitrary or disproportionate [001-96453].}
 \\ \bottomrule \bottomrule
\end{tabular}}
\caption{Case study 1 on a question from the AQuAECHR dataset, including provided source passages and model generations from Regular, AdaCAD, and CoCoLex.}
\label{tab_case1}
\end{table*}

\begin{table*}[h]
\centering
\scalebox{0.8}{
\begin{tabular}{p{1.5cm}p{18cm}}
\toprule
\toprule
Question & Under what circumstances can detention following conviction be considered unlawful?
\\
\midrule
Source Passages & 001-127697 56: The applicant alleged that what she considered to be the \hl{retroactive application} of a \hl{departure from the case-law} by the Supreme Court after she had been convicted had extended her detention by almost nine years, in violation of \hl{Article 7 of the Convention}, which reads as follows:

001-212829 35: In Kadusic v. Switzerland (no. 43977/13, 9 January 2018), which concerned a subsequent order of an institutional therapeutic measure under Swiss law, the Court was in principle prepared to accept that the order for an institutional measure amounted to a \hl{correction of the original judgment} following the discovery of \hl{relevant new circumstances} and that the fact that the measure was ordered in the context of proceedings for the review of a penalty imposed in a previous judgment may constitute a \hl{causal link between the initial conviction and the measure in issue}, as required by the relevant case-law of the Court concerning sub-paragraph (a) of \hl{Article 5 § 1} (ibid., § 50). However, in the circumstances of that case, the Court found that the detention which followed the (new) judgment in the review proceedings lacked a \hl{sufficient causal connection} with the initial conviction for being incompatible with its aims. The Court considered that the measure in question, which had been imposed a considerable period after the applicant’s initial conviction, had not been based on a sufficiently recent expert report and that the applicant had been detained in an institution unsuited to his mental disorders (ibid., §§ 53-60).

001-187391 116: The Court is therefore of the general view that in order not to make it impracticable for the police to fulfil their duties of maintaining order and protecting the public, provided that they comply with the underlying principle of Article 5, which is to \hl{protect the individual from arbitrariness} (see, Austin and Others v. the United Kingdom [GC], nos. 39692/09 and 2 others, § 56, ECHR 2012), the lawful detention of a person outside the context of criminal proceedings can, as a matter of principle, be permissible under \hl{Article 5 § 1 (c)} of the Convention. As regards the circumstances in which such detention is justified, this is an issue relating to compliance with Article 5 § 1 (c) and will be dealt with in paragraphs 143 to 174 below. \\ \midrule
\midrule
Regular & Detention following conviction can be considered unlawful under the principles outlined in ECHR case law if it lacks a \hl{sufficient causal connection} with the initial conviction [001-212829, § 53] or is based on an expert report that was not recent enough at the time of detainment. Additionally, such detentions must comply with Article 7 (prohibition against retroactive application) and \hl{protect individuals from arbitrariness} as required by the underlying principle of Article 5 to ensure their rights are protected during this period [001-127697].
 \\ \midrule
AdaCAD & Detention following conviction can be considered unlawful if there is a \hl{lack of sufficient causal connection} between the initial conviction and subsequent detension [001-212829, § 53]. Additionally, it may also occur when such detainment takes place in an institution that does not suit mental disorders present with the individual being held [ibid., §§ 60], or occurs considerable period after the applicant's original judgment \hl{without recent expert report}. \\\midrule
CoCoLex & Detention following conviction can be considered unlawful under the principles outlined in the case law if it lacks a \hl{sufficient causal connection} with the initial conviction [001-212829, § 50] or if it is based on a new judgment in the review proceedings that has been imposed a considerable period after the applicant's initial conviction and has not been based on a \hl{sufficiently recent expert report} [001-212829, §§ 53-60]. Additionally, detention outside the context of criminal proceedings can be unlawful if it is not permissible under \hl{Article 5 § 1 (c)} of the Convention and does not comply with the underlying principle of Article 5, which is to \hl{protect the individual from arbitrariness} [001-187391, § 56].
 \\ \bottomrule \bottomrule
\end{tabular}}
\caption{Case study 2 on a question from the AQuAECHR dataset, including provided source passages and model generations from Regular, AdaCAD, and CoCoLex.}
\label{tab_case2}
\end{table*}
\section{Case Study}
\label{case}
We provide Case Study 1 in Table \ref{tab_case1} and responses provided by three decoding methods—Regular, AdaCAD, and CoCoLex—to a legal question derived from the AQuAECHR dataset. The question at hand revolves around how the Court determines whether restrictions on the right to marry are properly regulated and subjected to judicial review and how this impacts an applicant's ability to challenge the denial of authorization. 

Among the methods, CoCoLex stands as the most faithful to the source document by directly incorporating key legal terminology and phrases such as "arbitrary or disproportionate interference", "necessity test", and "pressing social need." It maintains the original text's structure and legal nuances, ensuring an accurate replication of the Court's reasoning and intent, including the distinction that Article 12 does not require "necessity" or "pressing social need" tests, making it the most complete response in terms of correctness. 

On the other hand, Regular provides a structured response that correctly identifies the Court’s approach, emphasizing the need for restrictions to be grounded in concrete facts and national law. However, it does not explicitly mention the lack of "necessity" or "pressing social need" tests under Article 12, which is an important nuance. While this answer stays faithful to the source in terms of reasoning, it tends to oversimplify legal concepts, losing important subtleties and omitting essential legal terms that could distort the original meaning. The Regular answer is still well-organized and easy to follow, presenting the Court's reasoning in two key principles, but it generalizes the legal reasoning rather than explicitly referencing the differentiation between Article 12 and Article 8. 

AdaCAD captures the limitation imposed by national laws but is less precise in explaining the Court’s review process. The sentence structure makes it unclear how the judicial review process operates in practice. It deviates by implying that refusal could be seen as additional punishment without proper justification but does not clarify that the Court focuses on arbitrariness and proportionality rather than "necessity" or "social need.". It struggles with readability due to awkward phrasing and missing articles, which result in some unclear sentences. Though it attempts to reflect the source, its clarity is hindered by these issues.

We present Case Study 2 in Table \ref{tab_case2}, sourced from the AQuAECHR dataset, which explores the circumstances under which detention following conviction can be considered unlawful.

CoCoLex is the most faithful to the source document. It integrates key legal nuances, such as the specific conditions under Article 5 § 1 (c) of the Convention and the relationship between expert reports and the detention process. This response accurately reflects the structure and content of the legal text, ensuring a faithful representation of the original case law. It is comprehensive, addressing both the requirement of a causal connection to the initial conviction and the protection against arbitrariness under Article 5. The phrasing is formal yet accessible, offering a clear exposition of the legal conditions under which detention following conviction may be deemed unlawful. Its consistent use of legal terminology enhances both the fluency and clarity of the answer. The response is logically structured, presenting the necessary legal principles in a sequence that mirrors the original case law, starting with the causal connection to the initial conviction, followed by the relevance of expert reports and protection against arbitrariness. Each condition under which detention is unlawful is clearly linked, making the response coherent and well-organized.

Regular is relatively faithful but somewhat less precise. It correctly identifies the necessity of a causal connection with the initial conviction and emphasizes the protection against arbitrariness as required by Article 5. However, it oversimplifies the discussion of expert reports and the specific timeframe for lawful detention, omitting critical details such as the requirement for expert reports to be sufficiently recent and the need for expert reports in the context of a review. The response is coherent but lacks the tight organization found in CoCoLex. The sequence of ideas could be more logically presented, particularly regarding the role of expert reports and the protections under Article 5. While the phrasing is generally clear, it could benefit from more nuanced transitions to reflect the complexity of the legal principles discussed. Additionally, although the structure is grammatically sound and easy to follow, some of the phrasing feels formulaic, particularly in the transitions between legal concepts.

AdaCAD includes the core principles but lacks the depth and legal specificity present in CoCoLex. While it accurately mentions the need for a causal connection and references the issue of an unsuitable institution, it does not clarify that detention may be deemed unlawful due to insufficient expert reports and omits the protection against arbitrariness under Article 5. The response is less comprehensive in conveying the full scope of the legal reasoning outlined in the source passages. AdaCAD struggles with coherence, as the flow of ideas is interrupted by awkward phrasing. For instance, the transition between the causal connection and the discussion of the suitability of the institution could be smoother. Additionally, some constructions are grammatically clunky, such as "detention takes place in an institution that does not suit mental disorders present with the individual being held," which would benefit from clearer phrasing. These issues hinder the fluency and coherence of the response, making it more difficult to read.

Overall, maintaining fidelity to the original text is essential, especially in the legal domain, where even minor changes in wording can significantly impact the interpretation of rights and responsibilities. A model that deviates from legal fidelity risks presenting misleading information, which could distort legal reasoning or misrepresent the Court's stance. This is especially problematic in judicial contexts, where clarity, accuracy, and legal consistency are critical. CoCoLex's approach ensures that the model’s output adheres to established legal principles, preserving the integrity of the Court's decisions and offering the most reliable method for generating legally faithful responses.

\end{document}